\documentclass[10pt,twocolumn,letterpaper]{article}

\usepackage{cvpr}              

\definecolor{cvprblue}{rgb}{0.21,0.49,0.74}
\usepackage[pagebackref,breaklinks,colorlinks,allcolors=cvprblue]{hyperref}

\usepackage{color, xcolor} 
\usepackage{colortbl}
\definecolor{1}{RGB}{255,146,146}
\definecolor{2}{RGB}{255,204,153}
\definecolor{3}{RGB}{255,255,153}
\newcommand{\PreserveBackslash}[1]{\let\temp=\\#1\let\\=\temp}
\newcolumntype{C}[1]{>{\PreserveBackslash\centering}p{#1}}
\newcolumntype{R}[1]{>{\PreserveBackslash\raggedleft}p{#1}}
\newcolumntype{L}[1]{>{\PreserveBackslash\raggedright}p{#1}}
\usepackage{graphicx}
\usepackage{adjustbox}
\def\confName{CVPR}
\def\confYear{2025}

\title{SuperGS: Super-Resolution 3D Gaussian Splatting Enhanced by Variational Residual Features and Uncertainty-Augmented Learning}

\author{Shiyun Xie\\
Beihang University\\
Beijing, China\\
{\tt\small xieshiyun@buaa.edu.cn}
\and
Zhiru Wang\\
Beihang University\\
Beijing, China\\
{\tt\small 19241085@buaa.edu.cn}
\and
Xu Wang\\
Beihang University\\
Beijing, China\\
{\tt\small 21421012@buaa.edu.cn}
\and
Yinghao Zhu\\
Beihang University\\
Beijing, China\\
{\tt\small zhuyinghao@buaa.edu.cn}
\and
Chengwei Pan\thanks{Corresponding author.}\\
Beihang University\\
Beijing, China\\
{\tt\small pancw@buaa.edu.cn}
\and
Xiwang Dong\\
Beihang University\\
Beijing, China\\
{\tt\small xwdong@buaa.edu.cn}
}

\begin{document}
\maketitle
\begin{abstract}
Recently, 3D Gaussian Splatting (3DGS) has exceled in novel view synthesis (NVS) with its real-time rendering capabilities and superior quality. However, it faces challenges for high-resolution novel view synthesis (HRNVS) due to the coarse nature of primitives derived from low-resolution input views. To address this issue, we propose Super-Resolution 3DGS (SuperGS), which is an expansion of 3DGS designed with a two-stage coarse-to-fine training framework. In this framework, we use a latent feature field to represent the low-resolution scene, serving as both the initialization and foundational information for super-resolution optimization. Additionally, we introduce variational residual features to enhance high-resolution details, using their variance as uncertainty estimates to guide the densification process and loss computation. Furthermore, the introduction of a multi-view joint learning approach helps mitigate ambiguities caused by multi-view inconsistencies in the pseudo labels. Extensive experiments demonstrate that SuperGS surpasses state-of-the-art HRNVS methods on both real-world and synthetic datasets using only low-resolution inputs. Code is available at \url{https://github.com/SYXieee/SuperGS}.
\end{abstract}

\section{Introduction}
\label{sec:intro}

\begin{figure}[!ht]
\centering
\includegraphics[width=1.0\linewidth]{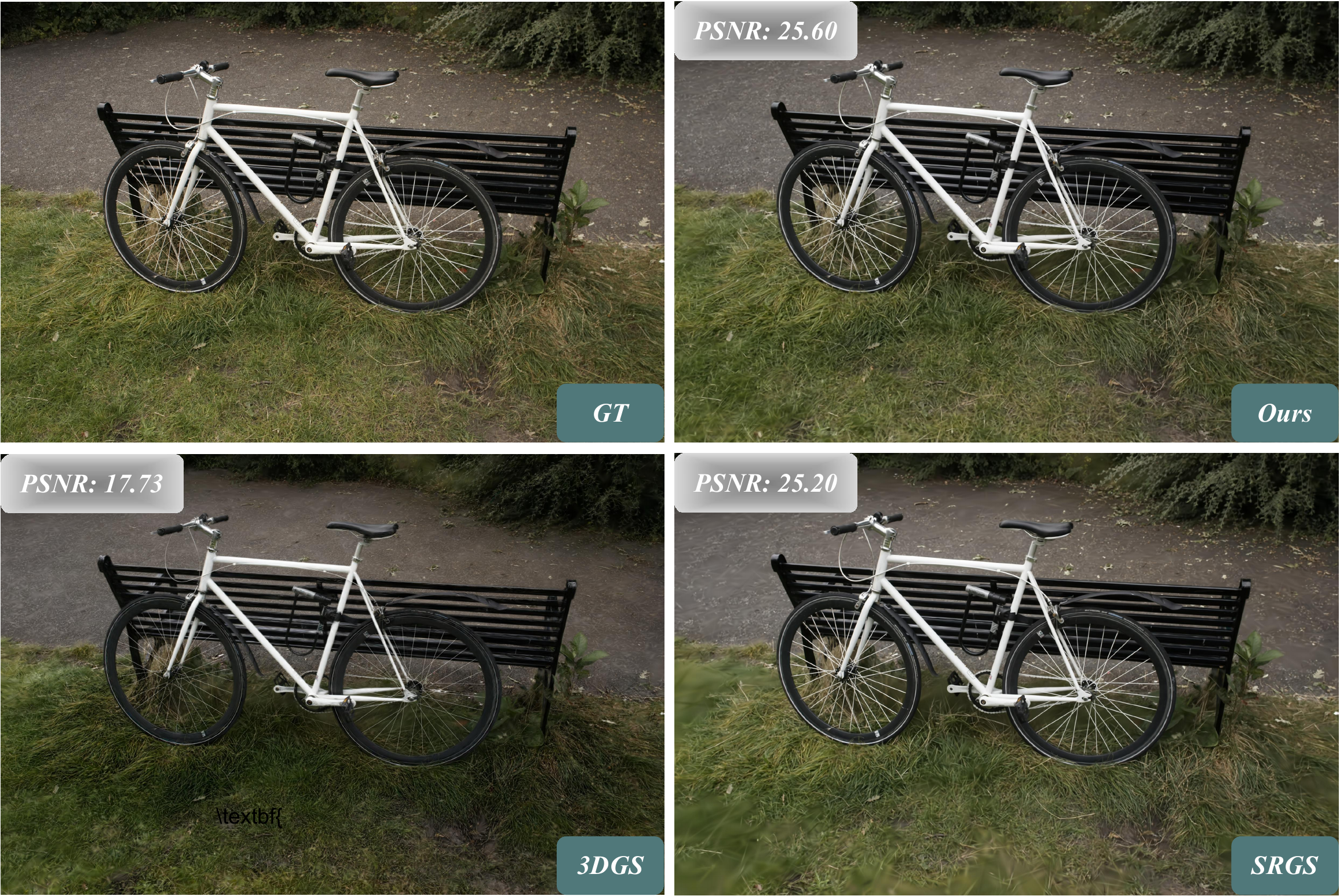}
\caption{\textit{Comparison of 3DGS and SRGS on HRNVS task.} 3DGS suffers from erosion effects, and SRGS exhibits significant artifacts, while Our method produces high-fidelity synthesized views with enhanced detail preservation.}
\label{fig:first}
\end{figure}

Novel view synthesis (NVS) is crucial for applications like AR/VR, autonomous navigation, and 3D content creation in the field of computer vision and graphics~\cite{deng2022fov, tonderski2024neurad, poole2022dreamfusion}. Traditional methods using meshes and points often sacrifice quality for speed~\cite{munkberg2022extracting, botsch2005high, yifan2019differentiable}. Conversely, Neural Radiance Fields (NeRF)~\cite{mildenhall2021nerf, barron2021mip, barron2022mip} have enhanced these tasks by implicitly modeling scene geometry and radiance, though their intensive computation limits real-time use~\cite{muller2022instant, fridovich2022plenoxels, Chen2022ECCV}. A promising approach is 3D Gaussian Splatting (3DGS)~\cite{kerbl20233d}, which uses 3D Gaussian primitives and a differentiable rasterization process for real-time, high-quality rendering. This technique avoids extensive ray sampling and employs clone and split strategies to enhance spatial coverage and scene representation. 

However, when handling high-resolution novel view synthesis (HRNVS), the vanilla 3DGS experiences significant performance degradation. Due to the lower sampling frequency of low-resolution images, primitives optimized in low-resolution scenarios often appear too coarse for high-resolution rendering, resulting in a loss of detail and visible artifacts~\cite{yu2024mip}. Unlike NeRF-style models, which can sample the color and opacity of any point due to their continuous implicit scene representation, 3DGS cannot directly upsample the Gaussian primitives. Existing methods~\cite{feng2024srgs, hu2024gaussiansr, shen2024supergaussian} typically use 2D priors to guide high-resolution scene optimization. However, 2D image super-resolution (SR) models upsample images individually, which can lead to multi-view inconsistencies. Furthermore, current video SR models do not yet possess the same powerful representational capabilities as image models.

To tackle these challenges, we propose a two-stage coarse-to-fine framework. In this framework, we first optimize the scene representation using low-resolution input views, which serves as the initialization for subsequent super-resolution. Moreover, the information learned at low resolution is retained as the foundational representation in the fine training stage, ensuring that the high-resolution optimization remains faithful to the original scene while focusing on enriching the details.

To extract additional high-resolution information, we leverage a pretrained Single Image Super-Resolution (SISR) model as a 2D prior to guide detail enhancement. To prevent training collapse caused by multi-view ambiguities from inconsistent pseudo labels, we introduce multi-view joint training, which focuses optimization on consistent regions while allowing gradients in inconsistent areas to cancel each other out, thus avoiding erroneous fitting. Additionally, we attach a learnable variational residual feature to each Gaussian primitive, which supplements high-resolution details while modeling uncertainty. High uncertainty typically indicates areas with large reconstruction errors or ambiguous supervision from different viewpoints. We then use this uncertainty to guide the densification process and loss computation, further improving the quality of the generated output, as shown in Figure~\ref{fig:first}.

In summary, we make the following contributions:
\begin{itemize}
    \item \textit{Insightfully}, we introduce SuperGS, an expansion of 3DGS for 3D scene super-resolution. SuperGS leverages a SISR model to guide detail enhancement while avoiding ambiguities caused by multi-view inconsistencies.
    \item \textit{Methodologically}, we propose a two-stage coarse-to-fine framework, which offers an optimal initialization and stores coarse information for super-resolution by constructing a feature field obtained from the low-resolution scene. We also introduce a variational residual feature that supplements high-resolution details while modeling uncertainty, further enhancing detail fitting and ambiguity mitigation.
    \item \textit{Experimentally}, extensive experiments demonstrate that SuperGS consistently surpasses state-of-the-art methods on both real-world and synthetic datasets.
\end{itemize}

\begin{figure*}[!ht]
\centering
\includegraphics[width=1.0\linewidth]{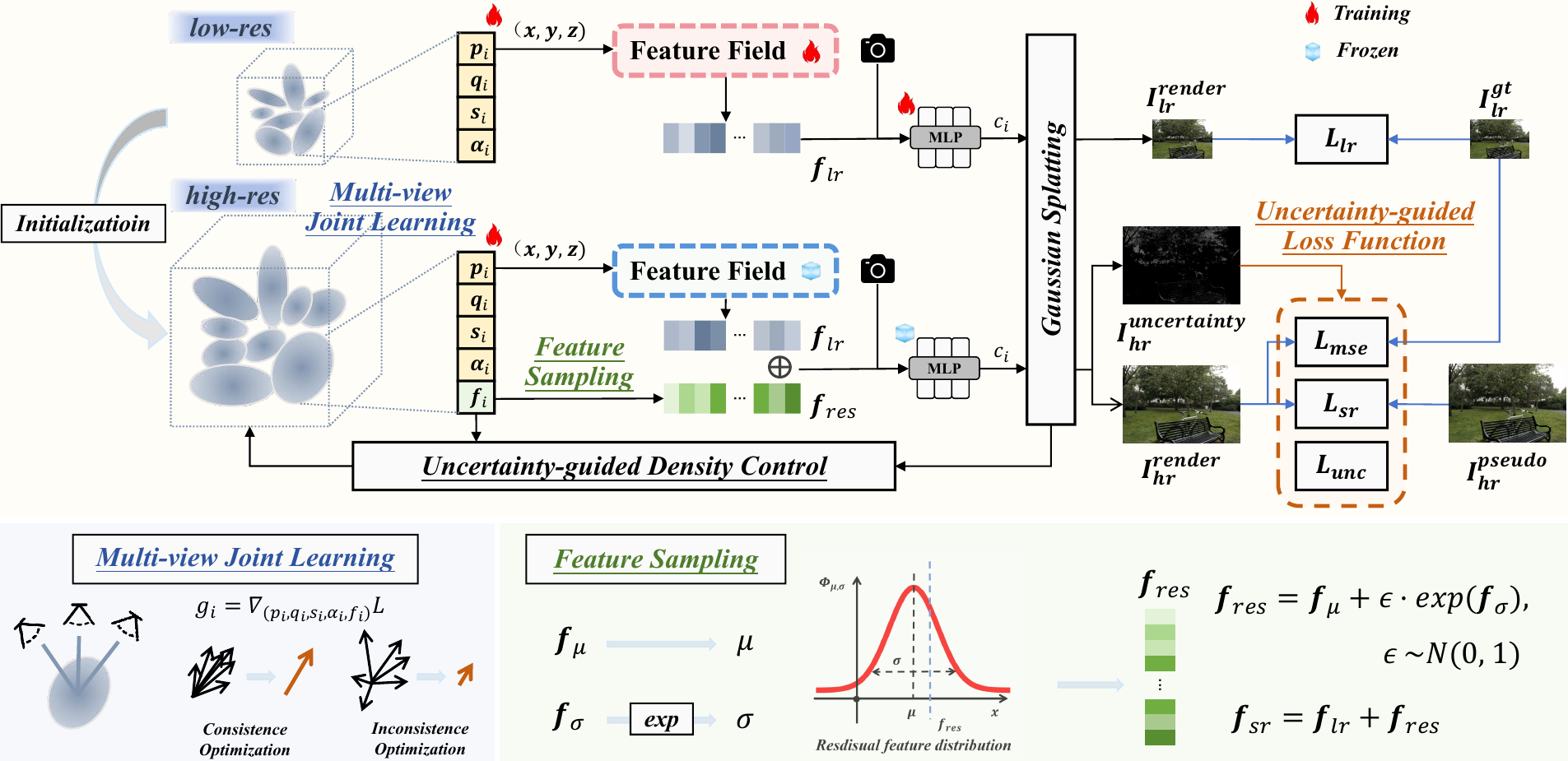}
\caption{\textit{Overview of our proposed SuperGS.} We propose a two-stage coarse-to-fine framework that first optimizes the scene with low-resolution views as initialization for super-resolution. In the coarse-stage, we introduce a latent feature field in place of the conventional 3DGS pipeline. In the fine-stage, we attach variational residual features to each Gaussian to enhance details and model uncertainty. Additionally, multi-view joint learning and uncertainty help mitigate ambiguities from pseudo labels, improving reconstruction quality.}
\label{fig:overview}
\end{figure*}

\section{Related Work}
\label{sec:formatting}

\subsection{Novel View Synthesis}

Novel view synthesis (NVS) is a complex task in computer vision, aimed at generating new viewpoint images from a set of posed photos. Neural radiance fields (NeRF)~\cite{mildenhall2021nerf} have excelled in NVS, encoding scene details into a MLP for volume rendering. However, NeRF's limitations in detail capture and efficiency led to the development of Mip-NeRF~\cite{barron2021mip} and InstantNGP~\cite{muller2022instant}, which enhance both aspects. Subsequent research~\cite{liu2020neural, hedman2021baking, lin2022efficient} has focused on finding a balance between rendering quality and efficiency, a persistently challenging task.

The recent development of 3D Gaussian Splatting (3DGS)~\cite{kerbl20233d} offers a potential solution to overcome the limitations of NeRF. By converting scene point clouds into anisotropic Gaussian primitives following Structure from Motion (SFM), 3DGS uses these primitives for explicit scene representation and employs a differentiable rasterizer for real-time, high-quality rendering. Due to its remarkable results, subsequent research has focused on improving the inherent limitations of 3DGS and applying its framework to various challenging tasks. Methods like AbsGS\cite{ye2024absgs} and Mini-splatting\cite{fang2024mini} address the shortcomings of 3DGS's densification strategy by proposing strategies that control Gaussian splitting or cloning based on gradient magnitude thresholds and the size of the view plane area influenced by dominant Gaussians. These strategies have significantly enhanced the rendering quality of 3DGS. SparseGS\cite{xiong2024sparsegs} tackles the challenging task of reconstruction under sparse viewpoints by integrating depth priors with constraints generated by denoising models and explicit constraints, thereby reducing background collapse and removing noise to produce robust 3D scene representations. SpecGS\cite{yang2024spec} improves upon 3DGS by replacing the spherical harmonics (SH) representation used for simulating view-dependent effects with a learnable anisotropic Gaussian (ASG) representation. By integrating this with an anchor method, SpecGS captures near-field lighting variations and more complex view-dependent effects with greater fidelity. Similarly, SpecLatentGS\cite{wang2024specgaussian} aims to enhance modeling of view-dependent effects by introducing latent features. It employs two parallel decoders to fit colors, achieving impressive results in complex lighting environments. ScaffoldGS\cite{lu2024scaffold} utilizes sparse voxel grids to initialize anchors that guide local 3D Gaussian distributions, forming a hierarchical and region-aware scene representation, resulting in more robust outcomes. Additionally, in recent years, 3DGS has been widely adopted and applied in fields such as virtual humans\cite{yuan2024gavatar}, dynamic scene reconstruction\cite{duan20244d,huang2024endo}, autonomous driving\cite{yan2024street}, and large-scale scene generation\cite{lin2024vastgaussian}.

\subsection{3D Scene Super-Resolution}
High-resolution novel view synthesis (HRNVS) aims to synthesize high-resolution novel views from low-resolution inputs. NeRF-SR~\cite{wang2022nerf} samples multiple rays per pixel, enhancing multi-view consistency at the sub-pixel level. However, its depth-guided refinement strategy necessitates a ground truth high-resolution image as input, which is often difficult to provide in many scenarios. The same challenge applies to RefSR-NeRF~\cite{huang2023refsr}. Additionally, NeRF suffers from slow rendering speeds, a problem that becomes even more pronounced in high-resolution scenes. FastSR-NeRF~\cite{lin2024fastsr} addresses this issue by introducing Random Patch Sampling, which not only improves rendering speed but also enhances output quality by increasing the diversity of image patches seen by the SR model during training. Moreover, pretrained 2D models can provide rich priors for 3D super-resolution tasks. \cite{yoon2023cross} uses Single Image Super-Resolution (SISR) to generate high-resolution details, employing cross-guided optimization to ensure multi-view consistency. DiSR-NeRF\cite{lee2024disr} ensures multi-view consistency through iterative 3D synchronization and builds up image details using Renoised Score Distillation.

Recently, 3DGS has become popular for its high-quality rendering and real-time rendering speeds. GaussianSR~\cite{hu2024gaussiansr} leverages off-the-shelf 2D diffusion priors and uses Score Dilation Sampling (SDS)~\cite{gou2021knowledge} to extract 3D information from 2D knowledge. SRGS~\cite{feng2024srgs}, on the other hand, uses high-resolution inputs from SISR as pseudo labels and introduces sub-pixel constraints as regularization. However, the pseudo labels provided by SISR models often suffer from multi-view inconsistencies, leading to ambiguities during training and errors in reconstruction results. SuperGaussian~\cite{shen2025supergaussian} attempts to address this by sampling a smooth trajectory from a 3D scene to generate video, and then using a pretrained video super-resolution model to produce supervision labels. Nevertheless, since images are much easier to obtain than videos, existing 2D image models still offer superior feature representation capabilities compared to video models. Therefore, our method continues to rely on a SISR model as the 2D prior and mitigates the ambiguities caused by multi-view inconsistencies in pseudo labels through uncertainty modeling and multi-view joint learning. Additionally, the proposed feature field provides essential information for high-resolution scenes, ensuring that the optimization remains close to the ground truth.

\section{Methodology}
\label{method}

In this work, we introduce a Super-Resolution 3D Gaussian Splatting method for high-resolution novel view synthesis (HRNVS) from low-resolution input views, utilizing a two-stage coarse-to-fine training framework. Figure~\ref{fig:overview} provides an overview of our SuperGS pipeline. In the coarse-stage training, we obtain a rough scene representation using low-resolution ground-truth images, while in the fine-stage training, we enhance the details by leveraging high-resolution pseudo labels generated from a pretrained Single Image Super-Resolution (SISR) model. The two-stage process is described as follows.

\subsection{Coarse-stage Training}

During the coarse training stage, our goal is to establish a latent feature field that represents the low-resolution 3D scene, which serves as the foundational feature for the high-resolution representation. Note that the parameters of this latent feature field are frozen during the fine training stage. Thanks to this approach, in the subsequent fine-stage training, we not only provide a good initial coarse feature for the high-resolution Gaussian primitives but also ensure that the additional details do not deviate from the core scene information, mitigating ambiguities caused by multi-view inconsistencies in the pseudo labels.

A straightforward way to construct the feature field is using MLPs to map Gaussian positions and view directions into the latent space, but this has significant drawbacks. Specifically, shallow MLPs often fail to capture enough information from positional encodings, whereas deeper MLPs, despite their higher capacity, markedly increase computational costs and extend rendering times.

\begin{figure}[!ht]
\centering
\includegraphics[width=0.9\linewidth]{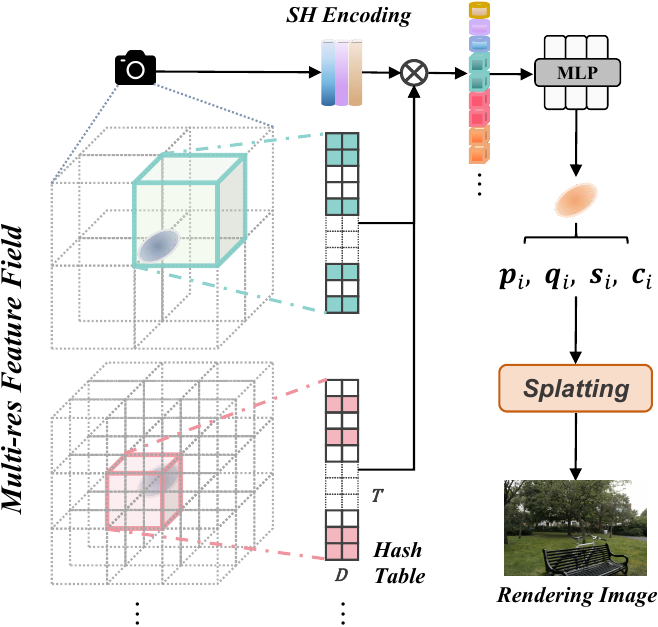}
\caption{\textit{Illustration of Feature Field.} For a specific Gaussian, we identify its voxel across $L$ resolution levels, extracting feature vectors of all voxel vertices from the corresponding hash table and deriving the $l$-th level feature via linear interpolation based on the Gaussian's center. We concatenate these features with SH encoding of direction, and a small MLP is used to decode the view-dependent color, which is finally rendered into an RGB image through a differentiable rasterizer.}
\label{fig:feature_field}
\end{figure}

Drawing inspiration from InstantNGP~\cite{muller2022instant}, we adopt a similar strategy by storing view-independent features in multi-resolution grids. This approach allows for efficient access to Gaussian features at arbitrary positions by indexing through hash tables followed by linear interpolation. As illustrated in Figure~\ref{fig:feature_field}, for each Gaussian, we retrieve learnable feature vectors stored at grids with varying resolutions and perform linear interpolation to derive its feature representation. It is important to note that different levels of grids correspond to distinct hash tables. Specifically, there are $L$ levels of grid resolutions, and the feature vector interpolated for the $i$-th Gaussian at the $l$-th level is denoted as $f_i^l$. From this, we construct its multi-resolution feature vector representation as follows:
\begin{equation}
    f_{lr_i} = f_i^1 \otimes f_i^2 \otimes ... \otimes f_i^L
\end{equation}
where $\otimes$ denotes the concatenation operation, and $f_{lr_i}$ is referred to as a view-independent feature because it is exclusively related to the coordinates of the Gaussian.

Similarly, to enable the grid feature field to extend to unbounded scenes, we reference the approach used in Mip-NeRF360~\cite{barron2021mip}, which normalizes the coordinates of the Gaussians into a contracted space. The contraction is formally described as follows:
\begin{equation}
    \mathrm{contract}(p_i) = \left\{\begin{array}{ll}
p_i, & ||p_i|| \le 1 \\
(2-\frac{1}{||p_i||})(\frac{p_i}{||p_i||}), & ||p_i|| > 1
    \end{array}\right.
\end{equation}
where $p_i \in \mathbb{R}^3$ represents the position of $i$-th Gaussian.

Additionally, to account for view-dependent information, we encode the view directions using Spherical Harmonics (SH) function, which is more suitable than component-wise frequency encoding. Subsequently, a tiny MLP (comprising two layers in practice) is used to map the concatenated feature vector and SH encodings to the final view-dependent color. This process is formally expressed as follows:
\begin{equation}
    c_i = \mathrm{MLP}(f_{lr_i} \otimes \mathrm{SH}(d_i))
\end{equation}
where $d_i \in \mathbb{R}^3$ denotes the view direction of the $i$-th Gaussian. It is noteworthy that, unlike binding SH parameters to each Gaussian, the same SH encoding and MLP parameters are used across the entire scene. Therefore, we just optimize a latent feature field that represents the entire scene instead of the SH parameters for each individual Gaussian.

After obtaining the color for each Gaussian primitive, we use the rendering pipeline from 3DGS~\cite{kerbl20233d} to generate the RGB image. By constructing the latent feature field, we overcome the feature discretization issues inherent in 3D Gaussian Splatting, allowing for feature retrieval at any location, which provides a solid foundation for subsequent high-resolution optimization.

\subsection{Fine-stage Training}

\subsubsection{Variational Residual Feature}

Lacking high-resolution detail, we utilize a pretrained SISR model to generate high-resolution input views as pseudo labels to guide detail enhancement. During the fine training stage, we freeze the parameters of the feature field, namely the hash table. We finetune the attributes of Gaussian primitives, including positions $p \in \mathbb{R}^3$, scales $s \in \mathbb{R}^3$, rotations $q \in \mathbb{R}^4$, and opacities $\alpha \in \mathbb{R}^1$. Additionally, we attach a residual feature vector to each primitive to supplement the details. To model uncertainty for each Gaussian, we assume that the residual features follow a normal distribution $\mathcal{N}(f_\mu, \exp(f_\sigma))$, learning both the mean and variance, where the exponential operation ensures the variance is non-negative. In total, the $i$-th Gaussian primitive during fine-stage training can be represented as:
\begin{equation}
    \mathcal{G}_i={(p_i, s_i, r_i, \alpha_i, f_{\mu_i}, f_{\sigma_i})}
\end{equation}

To achieve forward propagation and ensure gradients can backpropagate, we use the reparameterization trick~\cite{kingma2013auto} for residual feature sampling:

\begin{equation}
    f_{res_i}=f_{\mu_i}+{\epsilon}\cdot f_{\sigma_i}, \quad \epsilon \sim \mathcal{N} \mathcal(0, 1)
\end{equation}

The view-independent feature vector of the Gaussian is then represented as:
\begin{equation}
    f_{sr_i}=f_{lr_i}+f_{res_i}
\end{equation}
where $f_{lr_i}$ is retrieved and interpolated from the hash grid feature based on the position, and $f_{sr_i}$ represents the feature containing high-resolution details. This is then concatenated with the view encoding and decoded to obtain the view-dependent color.

Intuitively, $f_\mu$ and $f_\sigma$ model the distribution of the residual feature. In particular, the variance $f_\sigma$ encapsulates the uncertainty of the Gaussian, which may indicate that it fails to fit the reconstruction, represents floaters, or reflects inconsistencies in the multi-view pseudo labels. The residual feature $f_{res}$, on the other hand, represents a specific sample from this distribution.

\subsubsection{Multi-view Joint Learning}

MVGS~\cite{du2024mvgsmultiviewregulatedgaussiansplatting} uses multi-view training to mitigate overfitting to a specific view. In this paper, we also adopt the multi-view joint learning approach. Specifically, during each iteration, we sum the gradients from $M$ randomly selected view labels:
\begin{equation}
    g=\sum_i^M{g_i}
\end{equation}

As shown in Figure~\ref{fig:overview}, supervising with multiple pseudo labels simultaneously results in larger gradients in regions where the labels are consistent. In regions with higher inconsistencies, gradients either accumulate less, or when the gradient directions deviate significantly, they cancel each other out. This slows down parameter updates, helping to avoid fitting to a certain view and reducing ambiguity.

\begin{table*}[!ht]
        \footnotesize
        \centering
        \caption{\textit{Quantitative comparison for HRNVS ($\times 4$) on real-world datasets.} Mip-NeRF360: $1/8\rightarrow 1/2$, Deep Blending: $1/4\rightarrow 1/1$, Tansk\&Temples: $480\times270\rightarrow 1920\times1080$.}
        \label{tab:real_world}
    \begin{tabular}{L{3cm}|C{1cm}C{1cm}C{1cm}|C{1cm}C{1cm}C{1cm}|C{1cm}C{1cm}C{1cm}} 
\toprule
Dataset     & \multicolumn{3}{c|}{Mip-NeRF360} & \multicolumn{3}{c|}{Deep Blending} & \multicolumn{3}{c}{Tanks\&Temples}   \\
Method\&Metric      & PSNR$\uparrow$   & SSIM$\uparrow$  & LPIPS$\downarrow$   & PSNR$\uparrow$   & SSIM$\uparrow$  & LPIPS$\downarrow$  & PSNR$\uparrow$   & SSIM$\uparrow$  & LPIPS$\downarrow$ \\ \midrule
3DGS            &20.72 &0.617 &0.396
                &26.71 &0.839 &0.324
                &18.32 &0.612 &0.415 \\

3DGS-SwinIR     &26.15 &0.739 &0.316
                &28.36 &0.872 &0.306
                &20.50 &0.667 &0.396 \\

Mip-splatting   &26.42 &0.756 &0.294
                &28.85 &0.884 &0.282
                &20.65 &0.672 &0.385 \\
                
GaussianSR      &25.60 &0.663 &0.368
                &28.28 &0.873 &0.307
                &- &- &- \\
                
SRGS            &26.88 &0.767 &0.286
                &29.40 &0.894 &0.272
                &20.70 &0.687 &0.379 \\            
                \midrule
              
\textbf{Ours}   &\textbf{27.12} &\textbf{0.768} &\textbf{0.262} 
                &\textbf{29.77} &\textbf{0.899} &\textbf{0.268}
                &\textbf{21.19} &\textbf{0.695} &\textbf{0.364}  \\

\bottomrule
\end{tabular}
\end{table*}

\subsubsection{Uncertainty-guided Density Control}

We define the uncertainty $u_i$ of $i$-th Gaussian using the magnitude of its residual feature variance:
\begin{equation}
    u_i=||f_{\sigma _i}||_2
\end{equation}

Intuitively, high uncertainty indicates that the Gaussian either fails to fit the geometry of the region or represents floaters. Therefore, we take two actions for Gaussians with uncertainty exceeding a certain threshold.

First, noting that low-resolution primitives often appear too coarse for high-resolution rendering, and recognizing that high-resolution scenes require smaller Gaussians to capture finer details, we propose a splitting strategy based on both gradient and uncertainty. Formally, a Gaussian $\mathcal{G}_i$ is split once the following criteria are met:
\begin{equation}
(\nabla_{p_i} \mathcal{L} > \tau_p \text{ or } u_i > \tau_u) \text{ and } ||s_i|| > \tau_S,
\end{equation}
where $\tau_p$, $\tau_u$, and $\tau_s$ represent the gradient threshold, uncertainty threshold, and scale threshold, respectively.

Second, primitives with high uncertainty and low opacity are likely to be floaters, which may introduce visual artifacts and redundancy, reducing the quality of the final output. Therefore, we regularly remove these primitives to ensure a cleaner and more accurate scene representation.

\subsubsection{Uncertainty-guided Loss Function}

In addition to representing reconstruction errors, the Gaussian uncertainty also reflects the degree of multi-view inconsistency in the pseudo labels. When the inconsistency is high, the learned feature distribution tends to be more dispersed, i.e., the variance is larger. Therefore, in regions with high uncertainty, the influence of pseudo labels should be weakened, making the model more faithful to the low-resolution ground truth images. We achieve this by rendering an uncertainty map and incorporating it into the loss calculation.

To render the uncertainty map for a specific view, we adopt a similar point-based $\alpha$-blending approach. Given a camera pose $v$, we compute the uncertainty feature $U_{v,p}$ for a pixel by blending a set of ordered Gaussians that overlap with the pixel. The formula for $U_{v,p}$ is:

\begin{equation}
U_{v,p} = \sum_{i \in N} u_i a_i \prod_{j=1}^{i-1} (1 - a_j),
\end{equation}

where $\alpha$ is derived by evaluating a 2D Gaussian and is further adjusted by a learned opacity specific to each Gaussian.

We define the weighted $\mathcal{L}_1$, $\mathcal{L}_{MSE}$, and $\mathcal{L}_{SSIM}$ losses as follows:
\begin{equation}
    \begin{aligned}
        \mathcal{L}_1(I_x, I_y, w_p) &= \frac{1}{HW} \sum_p w_p |I_x(p) - I_y(p)| \\
\mathcal{L}_{MSE}(I_x, I_y, w_p) &= \frac{1}{HW} \sum_p w_p (I_x(p) - I_y(p))^2 \\
\mathcal{L}_{SSIM}(I_x, I_y, w_p) &= \frac{1}{HW} \sum_p w_p \cdot (1 - SSIM(I_x(p), I_y(p)))
    \end{aligned}
\end{equation}
where $w_p$ is the weight for pixel $p$, and $H$ and $W$ are the height and width of the image, respectively.

Based on this, we adaptively adjust the supervision strength of pseudo labels using the uncertainty map:
\begin{equation}
    \begin{aligned}
    \mathcal{L}_{sr} &= \mathcal{L}_1(I_{sr}^{render}, I_{sr}^{pseudo}, (1 - \sigma(U))) \\
    &+ \mathcal{L}_{SSIM}(I_{sr}^{render}, I_{sr}^{pseudo}, (1 - \sigma(U)))
    \end{aligned}
\end{equation}
where $\sigma$ is the sigmoid function.

At the same time, we regulate the high-uncertainty regions using the low-resolution ground truth images:
\begin{equation}
    \mathcal{L}_{mse} = \mathcal{L}_{MSE}(I_{lr}^{render}, I_{lr}^{gt}, \sigma(U))
\end{equation}
where $I_{lr}^{render}$ is obtained by downsampling $I_{hr}^{render}$ using average pooling.

In addition, we aim to minimize the uncertainty as much as possible:
\begin{equation}
    \mathcal{L}_{\text{unc}} = \frac{1}{HW} \sum_{p}U(p)
\end{equation}

Thus, the total loss during the fine-stage is defined as:

\begin{equation}
    \mathcal{L} = \mathcal{L}_{sr} + \alpha \mathcal{L}_{mse} + \beta \mathcal{L}_{unc}
\end{equation}
where $\alpha$ and $\beta$ are hyperparameters.

\begin{figure*}[!ht]
\centering
\includegraphics[width=0.95\linewidth]{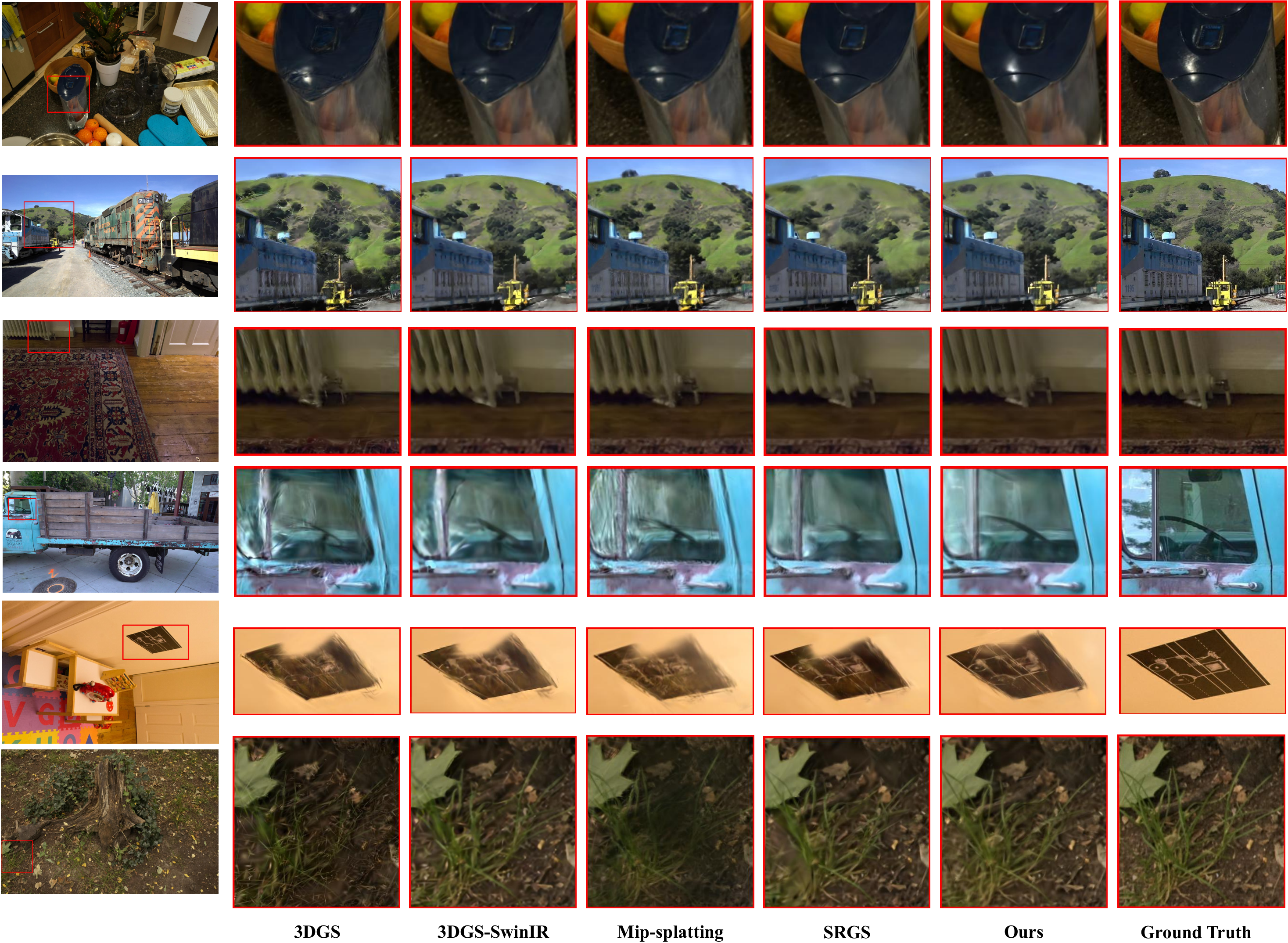}
\caption{\textit{Qualitative comparison of the HRNVS ($\times 4$) on real-world datasets.} We highlight the difference with colored patches.}
\label{fig:real_world}
\end{figure*}

\begin{figure*}[!ht]
\centering
\includegraphics[width=0.95\linewidth]{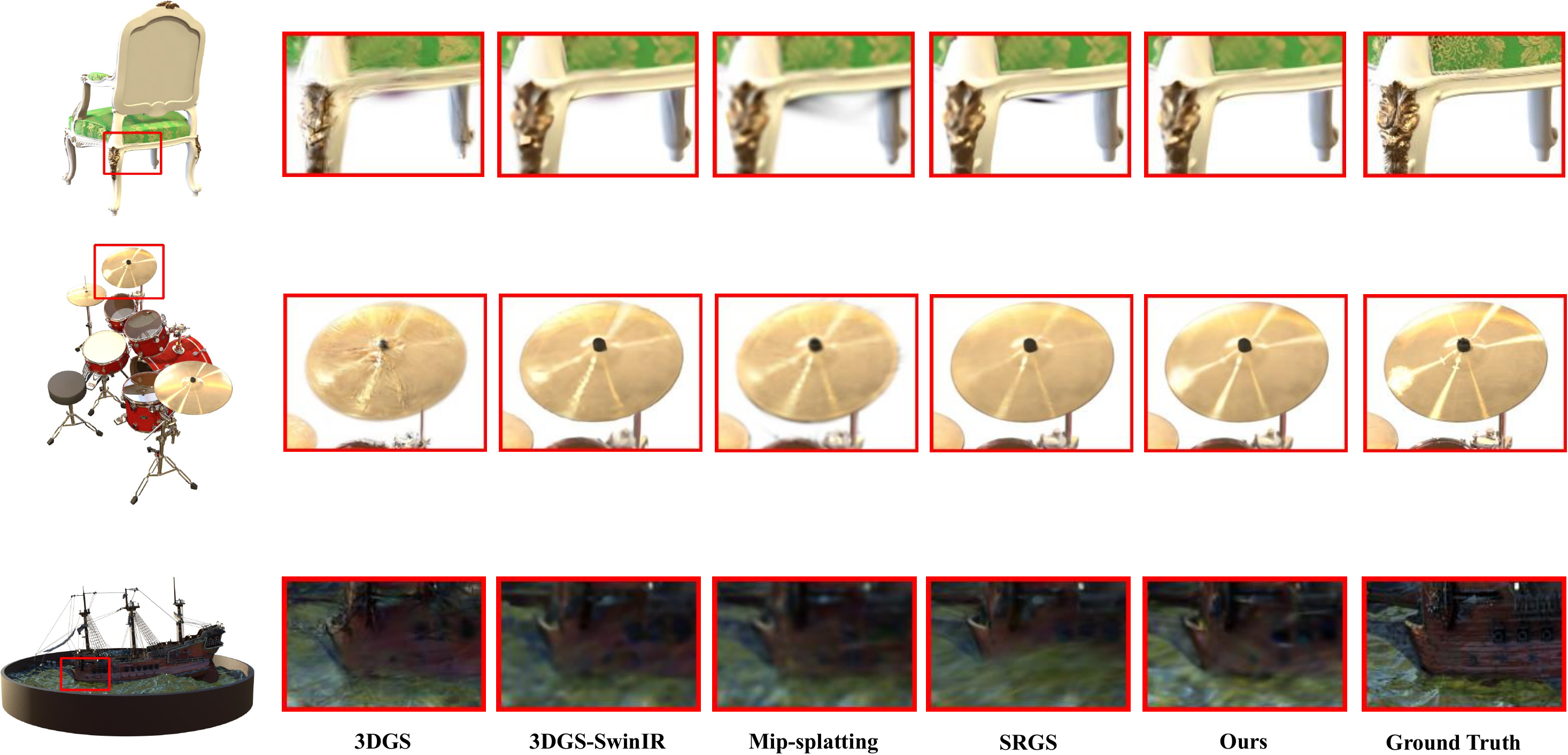}
\caption{\textit{Qualitative comparison of the HRNVS ($\times 4$) the synthetic dataset.} We highlight the difference with colored patches.}
\label{fig:blender}
\end{figure*}

\section{Experiments}

\subsection{Experimental Setups}

\subsubsection{Datasets and Metrics}
We conduct a comprehensive evaluation of our proposed model using PSNR, SSIM and LPIPS~\cite{zhang2018unreasonable} metrics. Our evaluation covers 21 scenes from both real-world and synthetic datasets, including 9 scenes from Mip-NeRF360~\cite{barron2022mip}, 2 scenes from Deep Blending~\cite{hedman2018deep}, 2 scenes from Tanks\&Temples~\cite{knapitsch2017tanks} and 8 scenes from NeRF Synthetic~\cite{mildenhall2021nerf}. For Mip-NeRF360, We downsample the training views by a factor of 8 as low-resolution inputs for the $\times 4$ HRNVS tasks. For Deep Blending, Tanks\&Temples and NeRF-Synthetic, we downsample the training views by a factor of 4 as inputs. 

\subsubsection{Baselines}
To validate the effectiveness of our method, we conduct comparisons with several prominent existing approaches. For 3DGS~\cite{kerbl20233d}, we train using low-resolution input views and directly render high-resolution images. Additionally, to maintain fairness in the comparisons, we utilize the same pretrained SISR model, SwinIR~\cite{liang2021swinir}, to upscale the low-resolution images rendered by 3DGS, referring to this as 3DGS-SwinIR. For Mip-splatting~\cite{yu2024mip}, which is a full-scale NVS method, we run the source code with its multi-resolution training setup. Regarding NeRF-SR~\cite{wang2022nerf}, FastSR-NeRF~\cite{lin2024fastsr}, GaussianSR~\cite{hu2024gaussiansr} and SuperGaussian~\cite{shen2025supergaussian}, we directly cite the results from respective papers under the same settings. And for SRGS~\cite{feng2024srgs}, we run the source code to obtain both qualitative and quantitative under identical conditions.

\subsubsection{Implementation Details}
We implement our framework based on the 3DGS~\cite{kerbl20233d} source code and modify the differentiable Gaussian rasterization to include uncertainty rendering. For a fair comparison, we also choose SwinIR~\cite{liang2021swinir} as the pretrained SISR model, mirroring SRGS~\cite{feng2024srgs}. For the latent feature field, we use 16 hash tables to store features at different levels, with each level containing 2-dimensional features. For the variational residual feature, to align with the hash grid features, we learn the mean $f_\mu \in \mathbb{R}^{32}$ and variance $f_\sigma \in \mathbb{R}^{32}$ separately. For multi-view joint learning, the number of views $M$ is set to 4 and 10 for real-world and synthetic datasets, respectively. For uncertainty-guided densification, we follow the gradient and scale thresholds of 3DGS, with $\tau_p$ and $\tau_s$ set to 0.00025 and 0.01, respectively, and set the uncertainty threshold $\tau_u$ to 0.5. For the uncertainty-guided loss function, the loss weights $\alpha$ and $\beta$ are set to 0.2 and 0.5, respectively. All our experiments are conducted on a single RTX A6000 GPU.

\subsection{Experimental Results}

Table~\ref{tab:real_world} and Table~\ref{tab:blender} showcase quantitative comparison results for the HRNVS tasks on the real-world and synthetic datasets. Our proposed SuperGS consistently and significantly outperforms previous state-of-the-art methods in terms of all three metrics across both indoor and outdoor scenes. As shown in Figure \ref{fig:real_world} and Figure~\ref{fig:blender}, 3DGS exhibits severe low-frequency artifacts due to the lack of high-resolution information, while Mip-splatting struggles with insufficient detail recovery due to missing high-frequency information. Although 3DGS-SwinIR enhances details, its separate upsampling of each view increases the risk of view inconsistency. SRGS also exhibits significant artifacts and reconstruction errors. In contrast, our method demonstrates superior precision and fidelity in detail recovery while substantially reducing artifacts that severely affect the visual quality of other methods.

\begin{table}[!ht]
        \footnotesize
        \centering
        \caption{\textit{Quantitative comparison for HRNVS ($\times 4$) on NeRF Synthetic dataset.} NeRF Synthetic: $200\times200\rightarrow 800\times800$.}
        \label{tab:blender}
    \begin{tabular}{L{2.2cm}|C{1cm}C{1cm}C{1cm}} 
\toprule
Dataset     & \multicolumn{3}{c}{NeRF Synthetic}   \\
Method\&Metric      & PSNR$\uparrow$   & SSIM$\uparrow$  & LPIPS$\downarrow$   \\ \midrule

3DGS            &21.77 &0.867 &0.104 \\

3DGS-SwinIR     &28.82 &0.933 &0.073\\


NeRF-SR         &28.90 &0.927 &0.099 \\

FastSR-NeRF     &30.47 &0.944 &0.075 \\

Mip-splatting   &24.59 &0.909 &0.101 \\
                
GaussianSR      &28.37 &0.924 &0.087 \\

SuperGaussian    &28.44 &0.923 &0.067 \\
                
SRGS            &30.83 &0.948 &\textbf{0.056} \\           
                \midrule
              
\textbf{Ours}   &\textbf{30.89} &\textbf{0.949} &\textbf{0.056} \\
\bottomrule
\end{tabular}
\end{table}

\subsection{Ablation Studies}
As shown in Table~\ref{tab:ablation}, we conduct ablation experiments on the components proposed in SuperGS, individually demonstrating the effectiveness of each part. The residual feature and multi-view joint learning methods, in particular, significantly improve reconstruction quality. Notably, while higher resolutions typically require more Gaussian primitives to accurately reconstruct the scene, using gradients alone as a condition for densification does not achieve the desired quality improvement. This may be due to ambiguities in the pseudo labels, where incorrect pseudo labels fail to push the gradient of misaligned Gaussians above the threshold. Incorporating uncertainty as a condition ensures a more effective densification process. Additionally, optimizing the loss function using uncertainty further enhances the realism of the reconstruction.

\begin{table}[!ht]
        \footnotesize
        \centering
        \caption{\textit{Ablation study results for HRNVS ($\times 4$) on Tanks$\&$Temples dataset.}}
        \label{tab:ablation}
\resizebox{\linewidth}{!}{
    \begin{tabular}{L{4cm}|C{1cm}C{1cm}C{1cm}} 
\toprule
Dataset     & \multicolumn{3}{c}{Tanks$\&$Temples}   \\
Method\&Metric      & PSNR$\uparrow$   & SSIM$\uparrow$  & LPIPS$\downarrow$   \\ \midrule

3DGS                    &18.32 &0.613 &0.416 
\\

$+$ Feature field    &18.34 &0.614 &0.429 \\

$+$ Residual feature     &20.88 &0.686 &0.381 \\

$+$ Multi-view joint learning  &21.14 &0.692 &0.371 \\

$+$ Gradient densification         &21.07 &0.693 &0.368 \\

$+$ Uncertainty-guided densification   
                        &21.17 &0.693 &0.368 \\ \midrule
                
$+$ Uncertainty-guided loss \textbf{(Ours)}    
            &\textbf{21.19} &\textbf{0.695} &\textbf{0.364} \\

\bottomrule
\end{tabular}
}
\end{table}

\subsection{Further Analysis}
\begin{figure}[htbp]
\centering
\includegraphics[width=0.9\linewidth]{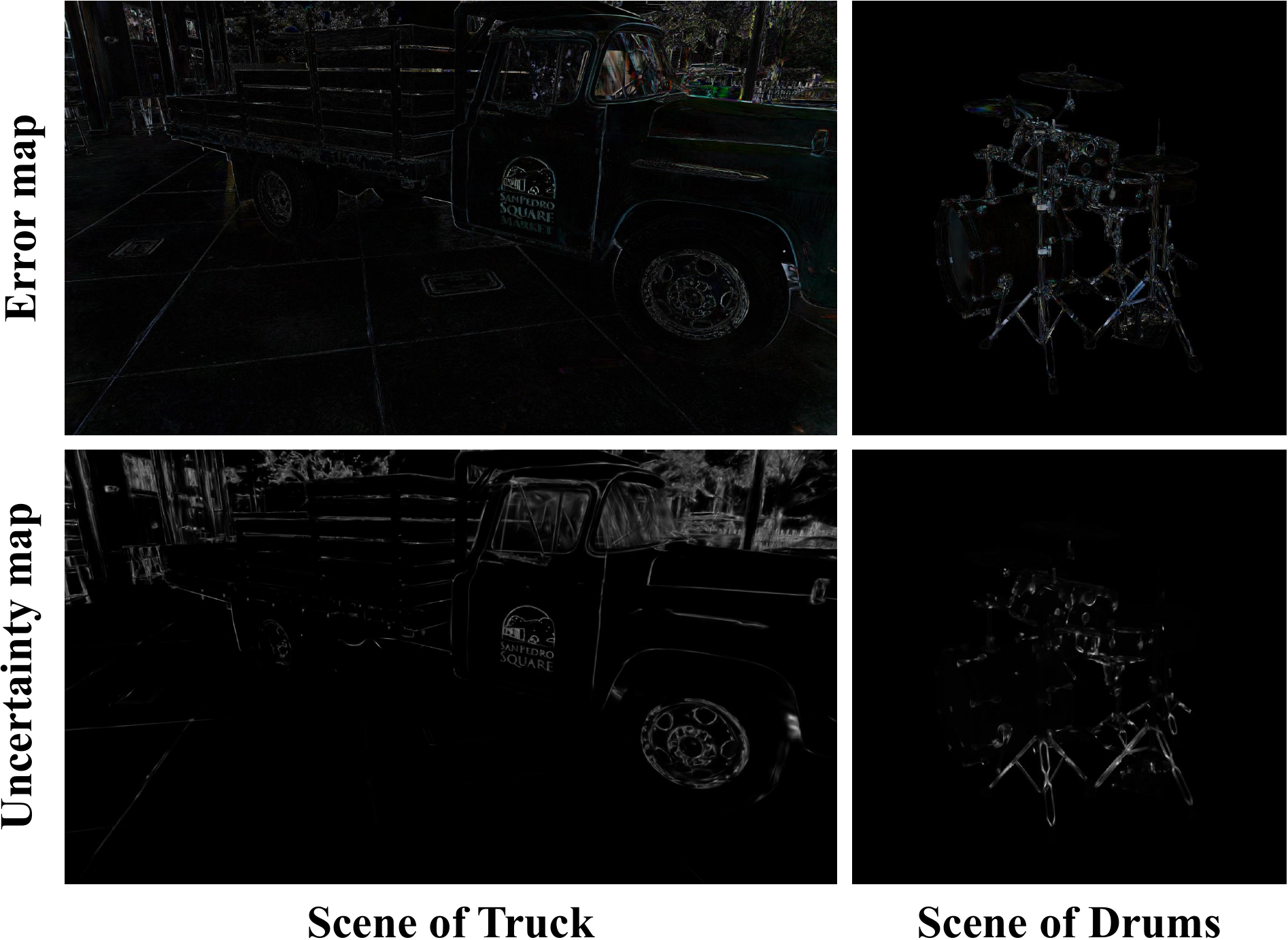}
\caption{\textit{Comparison between the uncertainty map and the error map.} A strong correlation between the two maps can be observed, indicating the validity of our approach to uncertainty modeling.}
\label{fig:uncertainty}
\end{figure}

We also visualize the rendered uncertainty map and compare it with the error map, which represents the difference between the rendered RGB image and the ground truth, as shown in Figure~\ref{fig:uncertainty}. We observe a significant overlap between areas of high uncertainty and regions with reconstruction errors, indicating a strong correlation between uncertainty and the accuracy of scene reconstruction. This further validates the effectiveness of our uncertainty modeling.

\section{Discussion}

\subsection{Limitations}

Although our method achieves significant results in the HRNVS task, it still faces the following limitations: 
\begin{itemize}
    \item The use of SISR models for pseudo label generation significantly increases training time costs, and the quality of pseudo labels largely determines the reconstruction quality of high-resolution scenes.
    \item The requirement for individual training for each scene, coupled with the lack of generalization capability, substantially restricts the range of applications.
\end{itemize}

\subsection{Future Work}
For future work, we plan to explore the development of a generalizable 3D super-resolution framework, which represents a promising direction given the powerful feature representation capabilities of current 2D image models. This approach not only has the potential to greatly reduce the training time costs associated with optimizing individual scenes but also holds promise for reducing image quantity requirements and expanding the scope of applications.

\section{Conclusion}

In this paper, we introduce SuperGS, an expansion of 3DGS designed to synthesize high-resolution novel views from only low-resolution inputs. The method is founded on a two-stage coarse-to-fine framework. In this framework, we first construct a coarse feature field and learn variational residual features to enhance details. Also, by introducing multi-view joint learning and modeling uncertainty, we significantly mitigate ambiguities arising from pseudo labels inconsistencies across multiple views. Comprehensive experiments on real-world and synthetic datasets demonstrate the effectiveness of our method in detail recovery and artifact elimination.


{
    \small
    \bibliographystyle{ieeenat_fullname}
    \bibliography{ref}
}


\end{document}